\documentclass[runningheads]{llncs}

\usepackage[T1]{fontenc}
\usepackage{amsfonts}
\usepackage{paralist}

\usepackage{tikz}
\usepackage{amsmath}
\usepackage{tabularray}
\usepackage{multirow}
\usepackage{adjustbox}
\usepackage{color}
\usepackage{graphicx,wrapfig,lipsum}
\usepackage{amssymb}
\usepackage[font=small,skip=3pt]{caption}
\usepackage[shortlabels]{enumitem}
\usepackage[ruled,vlined]{algorithm2e}
\usepackage{algpseudocode}
\usepackage{subcaption}
\usepackage{tcolorbox}

\usepackage[colorlinks,pagebackref,breaklinks]{hyperref} 

\hypersetup{
  hypertexnames=true, linkcolor=blue, anchorcolor=black,
  citecolor=magenta, urlcolor=alizarin  
}

\usepackage{color, colortbl}
\definecolor{Gray}{gray}{0.9}
\definecolor{airforceblue}{rgb}{0.36, 0.54, 0.66}
\definecolor{aliceblue}{rgb}{0.94, 0.97, 1.0}
\definecolor{alizarin}{rgb}{0.82, 0.1, 0.26}
\definecolor{amber}{rgb}{1.0, 0.75, 0.0}
\definecolor{amber(sae/ece)}{rgb}{1.0, 0.49, 0.0}
\definecolor{arsenic}{rgb}{0.23, 0.27, 0.29}
\definecolor{bronze}{rgb}{0.8, 0.5, 0.2}
\definecolor{battleshipgrey}{rgb}{0.52, 0.52, 0.51}
\definecolor{bole}{rgb}{0.47, 0.27, 0.23}
\definecolor{bulgarianrose}{rgb}{0.28, 0.02, 0.03}
\definecolor{cadet}{rgb}{0.33, 0.41, 0.47}
\definecolor{ceil}{rgb}{0.57, 0.63, 0.81}
\definecolor{cerulean}{rgb}{0.0, 0.48, 0.65}
\definecolor{charcoal}{rgb}{0.21, 0.27, 0.31}
\definecolor{coolblack}{rgb}{0.0, 0.18, 0.39}
\definecolor{coolgrey}{rgb}{0.55, 0.57, 0.67}
\definecolor{darkcandyapplered}{rgb}{0.64, 0.0, 0.0}
\definecolor{darkbrown}{rgb}{0.4, 0.26, 0.13}
\definecolor{darkcerulean}{rgb}{0.03, 0.27, 0.49}
\definecolor{darkgray}{rgb}{0.66, 0.66, 0.66}
\definecolor{darkjunglegreen}{rgb}{0.1, 0.14, 0.13}
\definecolor{darktaupe}{rgb}{0.28, 0.24, 0.2}
\definecolor{davy\'sgrey}{rgb}{0.33, 0.33, 0.33}
\definecolor{frenchblue}{rgb}{0.0, 0.45, 0.73}
\definecolor{almond}{rgb}{0.94, 0.87, 0.8}
\definecolor{beaublue}{rgb}{0.74, 0.83, 0.9}
\definecolor{beige}{rgb}{0.96, 0.96, 0.86}
\definecolor{bisque}{rgb}{1.0, 0.89, 0.77}
\definecolor{black}{rgb}{0.0, 0.0, 0.0}
\definecolor{fluorescentorange}{rgb}{1.0, 0.75, 0.0}
\definecolor{ghostwhite}{rgb}{0.97, 0.97, 1.0}
\definecolor{antiquewhite}{rgb}{0.98, 0.92, 0.84}

\makeatletter
\newcommand{\printfnsymbol}[1]{%
  \textsuperscript{\@fnsymbol{#1}}%
}
\makeatother

\begin{document}
%
\title{\centering To Vaccinate or not to Vaccinate? Analyzing $\mathbb{X}$ Power over the Pandemic}

\titlerunning{To Vaccinate or not to Vaccinate?}
\author{Tanveer Khan\inst{1}
\and Fahad Sohrab \inst{1} \and Antonis Michalas\inst{1, 2} \and Moncef Gabbouj \inst{1}}
 \institute{Tampere University, Finland \and RISE Research Institutes of Sweden \\
\email{(tanveer.khan, fahad.sohrab, antonios.michalas, moncef.gabbouj)@tuni.fi}
 }


\authorrunning{Khan et al.}


%
\maketitle              
\begin{abstract}
The COVID-19 pandemic has profoundly affected the normal course of life --  from lock-downs and virtual meetings to the unprecedentedly swift creation of vaccines. To halt the COVID-19 pandemic, the world has started preparing for the global vaccine roll-out. In an effort to navigate the immense volume of information about COVID-19, the public has turned to social networks. Among them, $\mathbb{X}$ (formerly Twitter) has played a key role in distributing related information. Most people are not trained to interpret medical research and remain skeptical about the efficacy of new vaccines. Measuring their reactions and perceptions is gaining significance in the fight against COVID-19. 
To assess the public perception regarding the COVID-19 vaccine, our work applies a sentiment analysis approach, using natural language processing of $\mathbb{X}$ data.  We show how to use textual analytics and textual data visualization to discover early insights (for example, by analyzing the most frequently used keywords and hashtags). Furthermore, we look at how people's sentiments vary across the countries. Our results indicate that although the overall reaction to the vaccine is positive, there are also negative sentiments associated with the tweets, especially when examined at the country level. Additionally, from the extracted tweets,  we manually labeled~100 tweets as positive and~100 tweets as negative and trained various One-Class Classifiers (OCCs). The experimental results indicate that the S-SVDD classifiers outperform other OCCs.

\keywords{COVID-19 \and One-Class Classifiers \and Twitter}

\end{abstract}

\section{Introduction}
\label{sec:introduction}
The year 2020 brought forth a new era for science.  The coronavirus pandemic (COVID-19) has affected every single aspect of our lives. The way we work and live has changed progressively -- virtual meetings, teleworking, and social distancing are the new norm of the day. Despite these challenges,~2020 also brought a remarkable advancement in science through an unprecedentedly fast development of new vaccines to combat the pandemic. From the onset of the pandemic, medical researchers have been trying to develop models to forecast the spread of the SARS-CoV-2 -- the virus causing COVID-19~\cite{cossarizza2020sars}. Lockdown surfaced as the main precautionary, non-pharmaceutical public health measure available to curb the spread of the virus. Many scientists believe that ending the COVID-19 pandemic can be achieved through herd immunity, i.e., a large proportion of the world population developing immunity to the virus -- a state achievable only through large vaccination campaigns~\cite{krammer2020sars}.

In less than a year from the COVID-19 outbreak, several pharmaceutical companies developed vaccines that offer varied levels of protection. First among those to make the headlines were Pfizer 
and Moderna. 
Many western countries have already embarked on large vaccination campaigns, with the promise of transparency, immunity against the virus and ultimately a return to normality. However, still, a significant number of people are against being vaccinated. In addition to that, a small yet fervid COVID-19 anti-vaccination movement is on the rise. Philip Ball in his report~\cite{ball2020anti} mentioned that, ``\textit{the anti-vaccine campaign might weaken the efforts to stop the COVID-19}''. Furthermore,  conspiracy theories' supporters are exploiting attention-grabbing statements such as ``\textit{vaccination will kill one million people}'' and ``\textit{through vaccination microchips will be implanted into people}'' and attracting increasing numbers of even the shrewdest of social media and internet users. 
Unfortunately, the rapid development of social networks~\cite{Michalas:2022:SecureComm:MetaPriv} makes the distribution of such news easier than ever. These platforms have grown into being one of the most widely used channels for the dissemination of information during the COVID-19 pandemic. Undoubtedly, the most popular social media platform for distributing news is $\mathbb{X}$. Over the years, $\mathbb{X}$ has become a crucial tool for journalists. It has spread rapidly through newsrooms and now plays a central role in the way stories are sourced, broken, and distributed -- contributing to a further acceleration of the news cycle. 
It provides a platform where people from all walks of life can connect, find, or share relevant and reliable information surrounding various topics, including the COVID-19 pandemic.  
$\mathbb{X}$ does not discriminate against COVID-19 vaccine supporters or deniers which makes it a suitable ground for the analysis of the public perception around the COVID-19 vaccine. Thus, in our work, we used it to identify tweets related to COVID-19 vaccination. 

\textbf{Contributions}: The main aim of our work was to perform sentiment analysis of a large number of tweets from users around the world and gauge the overall public perception towards COVID-19 and vaccines. Furthermore, we segmented the tweets based on the country of origin and correlated the sentiments of vaccine-related tweets with the country's situation. This helped us understand country-specific public perception towards vaccines. 
Our contribution can be summarized as follows:
\begin{enumerate}[1.]
	\item  First, using the $\mathbb{X}$ API Tweepy, we collected approximately~0.4 million tweets containing keywords concerning COVID-19 and COVID-19 vaccination. 
	\item Then, we conducted tweets'  textual analysis to identify the public sentiment. A sentiment score was assigned to each tweet using the Valence Aware Dictionary for Sentiment Reasoning (VADER). Additionally, the tweets were analysed to find the most popular keywords contained in them. 
	\item We also examined the tweets' global and country-level sentiments related to the pandemic and vaccines.
    \item We manually labeled 100 tweets as positive and 100 tweets as negative.
    \item We trained different OCCs on both the manually labeled positive and negative tweets. To assess the robustness of these models, we perform random selection five times, creating five different train-test splits for the experiments.
\end{enumerate}  

\section{Preliminaries}
\label{sec:preliminaries}
In one-class classification (OCC), the objective is to infer a data description only from a single data category \cite{sohrab2023graph}. The data used for creating a data description is termed the positive or target class, and all other categories are collectively referred to as the negative or outlier class \cite{sohrab2020ellipsoidal}. Over the past few decades, significant progress has been made in developing OCC algorithms, which have been applied across a variety of domains. For instance, in \cite{pantazi2019automated}, OCC is employed for automated leaf disease detection in different crop species. The use of OCC to enhance deep CNN-based taxa identification by flagging samples that may belong to rare classes for further human inspection is explored in \cite{sohrab2020boosting}. OCC methods have been employed to investigate early myocardial infarction detection \cite{zahid2024refining,degerli2022early}. The application of OCC in hyperspectral image analysis is showcased in \cite{kilickaya2023hyperspectral}. Moreover, \cite{guidry2023one} highlights its use in intrusion detection within vehicular networks, and \cite{zaffar2023credit} explores the effectiveness OCC methods in detecting credit card fraud.

One-Class Support Vector Machine (OC-SVM) \cite{scholkopfu1999sv} and Support Vector Data Description (SVDD) \cite{tax2004support} are the popular traditional OCC methods. OC-SVM constructs a hyperplane that separates the target class by maximizing its distance from the origin, effectively separating the target data. SVDD encapsulates the target class within a minimal-volume hypersphere in the given feature space, effectively modeling the class boundary by minimizing the sphere's volume around the target data. Numerous enhancements to OC-SVM and SVDD are proposed in the literature. In \cite{mygdalis2016graph}, a graph-embedded approach for one-class classification integrates a generic graph structure into the optimization framework of OC-SVM and SVDD, enhancing their adaptability. In \cite{sohrab2018subspace}, the subspace support vector data description (S-SVDD) method is introduced, focusing on optimizing subspace for one-class classification using an iterative gradient descent-based optimization process. Another variant of S-SVDD using Newton’s method (NS-SVDD) rather than gradient-based optimization is introduced in \cite{sohrab2023newton}. Unlike gradient descent, Newton’s method incorporates second-order information through Hessian matrix, enabling different optimization of objective.

Let the target class training samples, which are to be enclosed within a boundary, be represented by matrix $\mathbf{X}=[\mathbf{x}_{1},\mathbf{x}_{2},\dots, \mathbf{x}_{N}],\mathbf{x}_{i} \in \mathbb{R}^{D}$, where $N$ is number of samples and $D$ is data dimensionality. S-SVDD is formulated as:

\begin{equation}
\small
\min \quad F(R,\mathbf{a}) = R^2 + C\sum_{i=1}^{N} \xi_i
\ \ \textrm{s.t.} \quad  \|\mathbf{Qx}_i - \mathbf{a}\|_2^2 \le R^2 + \xi _i, \quad \xi_i \ge 0, \:\: \forall i\in\{1,\dots,N\}.
\end{equation}
Here, $R$ denotes radius, $\mathbf{a} \in \mathbb{R}^{D}$ is the center of hypersphere, and hyperparameter $C > 0$ regulates the balance between minimizing hypersphere's volume and permitting certain data points to lie outside the boundary. Slack variables $\xi_i, i=1, \dots, N$ are included to allow some target data to be outliers. The matrix $\mathbf{Q} \in \mathbb{R}^{d \times D}$ serves as the projection matrix, transforming data from the original $D$-dimensional feature space into an optimized, lower $d$-dimensional space. In S-SVDD, an augmented Lagrangian function with a regularization term, $\psi$, is optimized:
\begin{equation}\label{Lang}
L= \sum_{i=1}^{N} \alpha_i  \mathbf{x}_i^\intercal \mathbf{Q}^\intercal \mathbf{Q} \mathbf{x}_i - \sum_{i=1}^{N}\sum_{j=1}^{N} \alpha_i \mathbf{x}_i^\intercal \mathbf{Q}^\intercal \mathbf{Q} \mathbf{x}_j \alpha_j + \beta\psi,
\end{equation}
where $\alpha$ denotes the Lagrange multipliers, and $\beta$ is used to regulate the weight of the regularization term. The regularization term $\psi$ quantifies class variance: \begin{equation}
\label{generalconstraintpsi} 
\psi = \text{Tr}(\mathbf{Q}\mathbf{X}\boldsymbol{\lambda}\boldsymbol{\lambda}^\intercal 
 \mathbf{X}^\intercal\mathbf{Q}^\intercal),
\end{equation} 
where Tr($\cdot$) is the trace operator and $\lambda \in \mathbb{R}^N$ is a vector that selects the contribution of particular data points to the optimization process, leading to different variants of S-SVDD. These variants are as follows:

In the different variants of S-SVDD, the regularization term is utilized in distinct ways. In S-SVDD$\psi1$, the regularization term is excluded and does not contribute to the data description. In S-SVDD$\psi2$, all training samples are used to capture the class variance in the regularization term. S-SVDD$\psi3$ incorporates both the samples on the boundary and those outside it into the regularization term. Finally, in S-SVDD$\psi4$, only the support vectors at the class boundary are considered for describing the class variance in the regularization term. The projection matrix $\mathbf{Q}$ is updated through gradient- or Newton-based optimization techniques to iteratively refine the solution.

\section{Related work}
\label{sec:relatedwork}

Technology proved to be an indispensable tool in response to the COVID-19 pandemic. COVID-19-related information in 2020 was the highest-placed topic in social media. Inevitably, this also introduced a substantial amount of false information. The explosion of false information generating fear and panic among people worldwide\footnote{\url{https://shorturl.at/d6Kdy}}
was subsequently labeled as the first true social media "infodemic." In a study by Rani Molla\footnote{\url{https://shorturl.at/cM8xe}}, 
staggering~19 million online mentions of COVID-19 were recorded globally in only~24 hours. A media giant ABC News report stated that the fear of COVID-19 spread faster than the actual virus\footnote{\url{https://shorturl.at/AqmWj}}.

On the positive side, social media also provides an abundance of useful information related to COVID-19. The false information, however, in some instances, created uncertainty which had serious health consequences~\cite{khan2020trust,khan2021fake,khan2021seeing,khan2024trustworthiness}. A man reportedly died, and his wife was hospitalized due to drinking a form of Chloroquine believed to cure COVID-19\footnote{\url{https://shorturl.at/42fQK}}. 
Ahmad \textit{et al.,}~\cite{ahmad2020impact}, collected data from~516 users in order to examine how social media affects mental health as well as the spread of the COVID-19 related panic. Their results demonstrated that social media has a huge impact on spreading fear among people and a possible detrimental effect on their health. 

In a similar study, monitoring the public health concerns, Prabhsimran \textit{et al.,}~\cite{singh2020psychological} conducted sentiment analysis to determine the overall sentiment status of a $\mathbb{X}$ user. The experiments were carried out on a set of~10,403 tweets using the keywords COVID-19 and CORONAVIRUS. The study found that during the pandemic the negative sentiments in the tweets were predominant.
 
The lack of information related to COVID-19 caused misinformation to circulate at an alarming rate. Ramez \textit{et al.,}~\cite{kouzy2020coronavirus} focused on analyzing the magnitude of medical misinformation spread on $\mathbb{X}$. A total of~673 tweets were collected using~11 different hashtags and three popular keywords related to COVID-19. The considered tweets were in English and had been re-tweeted at least five times or more. According to the results, most of the tweets contained important content with information related to COVID-19. The findings were that the main topics of interest were public health, followed by socio-political and financial issues. The study also found that the hashtag ``\#COVID-19'' had the lowest while the hashtags ``\#2019\_ncov'' and ``\#Corona'' had the highest rate of misinformation. 

In another study, Bonnevie \textit{et al.,}~\cite{bonnevie2020quantifying} suggested that the negative criticism and deliberate spread of misinformation increased distrust among deniers of the COVID-19 vaccine. To ensure vaccine support and avoid significant health implications, the authors suggested it is crucial to separate accounts spreading reliable from those spreading unreliable information. In a similar study, Jamison \textit{et al.,}~\cite{jamison2020not} analyzed the content from the most active $\mathbb{X}$ accounts. Interestingly, they discovered that although vaccine deniers circulated misleading information, vaccine supporters also contributed to dissemination of non-reliable information. 

The increase in skepticism about the COVID-19 vaccines and social media posts from anti-vaccination campaigners are leading to vaccine hesitancy -- a phenomenon that makes the fight against the pandemic even more challenging. In an attempt to mitigate this, the World Health Organization report provided guidelines for healthcare workers on how to respond to vaccine deniers~\cite{schmid2018commentary}. The need for swift action on this level stresses the importance of timely management of this problem in order to reduce its negative impact. We believe that expeditiously and accurately identify the spread of misinformation can limit its impact and improve the functioning of our society. 

\section{Methodology}
\label{sec:methodology}

We focused on understanding people's perceptions regarding the COVID-19 vaccination. We harvested a large number of tweets and performed sentiment analysis to assess the public's view towards COVID-19 vaccination. We classified these views into two categories -- \textit{vaccine supporters} and \textit{vaccine deniers}. Data analysis to retrieve the most frequently used words, topic trends, etc.,  was conducted to gather insights and help us understand the support or suspicion for the COVID-19 vaccination campaign. This analysis is important as it could assist policymakers in checking public opinion and the current level of trust in the immunization campaign. Additionally, it can be used to assess the need for ramping up sensitization efforts when public perception leans more towards denying efficacy or safety of the vaccine.

To process and analyze the collected tweets, in order to measure the public reaction to the COVID-19 vaccine recommendation, we incorporated text mining methods using Natural Language Processing (NLP)~\cite{chowdhury2003natural}. We started by using Tweepy\footnote{\url{https://www.tweepy.org/}}  -- a popular $\mathbb{X}$ API -- to harvest tweets related to the recent vaccination campaign and created a dataset containing approximately~0.4 million tweets related to COVID-19 and COVID-19 vaccine. Next, we utilized NLP 
techniques to perform a sentiment analysis on the generated data. Sentiment analysis involves detecting the emotion of a text which can be either positive or negative. For our sentiment analysis, we used VADER 
and Textblob 
libraries. 

\begin{itemize}
    \item \textbf{Sentiment Analysis with VADER:} VADER is a library that is pre-trained using rule-based values. It is particularly attuned to sentiments expressed in social media. It can analyze a text and give evaluation not only related to the nature of the emotion, such as whether it is positive or negative, but also its strength. It uses a bag-of-words approach with simple heuristics (increasing sentiment intensity in the presence of certain words like ``really'' or ``very''). The reason for choosing this library is that it performs very well on social media text without requiring any training data.
    
    \item \textbf{Sentiment Analysis with TextBlob:} TextBlob is a Python library built upon a natural language toolkit used mainly for processing textual data. A variety of NLP tasks can be performed using this library. For a given input, it returns two values -- positive or negative polarity. Similar to VADER, TextBlob is also based on a bag-of-words approach. 
\end{itemize}
  

\begin{figure}[h]
	\centering
	\includegraphics[width=0.8\linewidth]{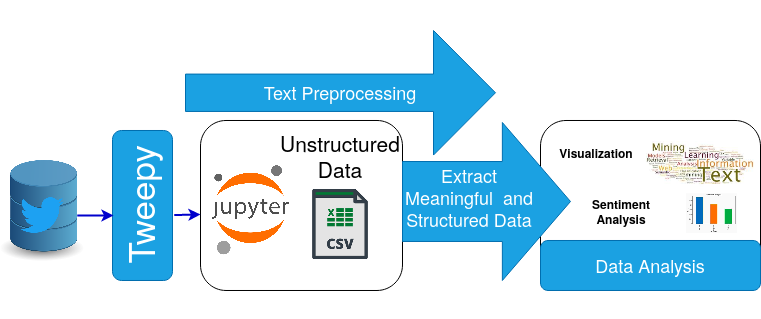}
	\caption{ Tweets' sentiment analysis flow}
	\label{fig:Proposed Protocol Overview}
\end{figure}

As previously mentioned, for the needs of this study, we generated a dataset by collecting $\mathbb{X}$ users' data using the Tweepy API. Our dataset contained tweets related to COVID-19 vaccination, identified using a list of associated hashtags. 
Identified hashtags were used to crawl $\mathbb{X}$'s posts and extract data containing one or more of these words. Using this list, we collected~0.4 million tweets from the year~2021. Starting with this dataset we subsequently excluded non-English characters, stop words, punctuation, special characters, and URL's. Further, we normalized the text by performing stemming -- removing unnecessary suffixes from beginning or end of the word and lemmatization -- analysing a word in terms of vocabulary. This allowed us to create a cleaner dataset since these words do \textit{not} contribute to the sentiment meaning of messages. In addition to that, from each tweet, we also extracted information such as username, location, hashtags, and date. While this data is important for a number of reasons (e.g., generating word cloud, trend analysis, identifying current trends by specifying the number of tweets within a specific timeline, etc.), for our sentiment analysis, the ``tweet text'' was the most critical information(~\autoref{fig:Proposed Protocol Overview}). Furthermore, from the collected tweets, we manually labeled~100 as positive and~100 as negative. Two experts in the field did this manual classification. Using this dataset, we trained different OCC models, including SVDD, ESVDD, OCSVM, S-SVDD$\psi_{1}$, S-SVDD$\psi_{2}$, S-SVDD$\psi_{3}$ and S-SVDD$\psi_{4}$. Unlike traditional binary classifiers, OCCs rely on training data from a single class, making them ideal when abundant data exist for one class but little to none for others. These OCCs were trained separately on both positive and negative tweets to infer the decision function for each class.

\textbf{\textit{Experimental Setup}}: To extract the features from $\mathbb{X}$ and generate the dataset, we used Python~3.5. The Python script was executed locally on a machine with the following configuration: Intel Core~i7,~2.80 GHz,~32GB, Ubuntu~16.04 LTS~64 bit. For sentiment analysis we used textblob\footnote{\url{https://textblob.readthedocs.io/en/dev/}} and vaderSentiment\footnote{\url{https://pypi.org/project/vaderSentiment/}} python libraries and the script was executed on the Google Colab.

For evaluating the performance of OCC methods, we selected 70\% of data for training and the remaining 30\% for testing while keeping the class proportions in both sets matched those in the full dataset. Each experiment was repeated five times with different random train/test splits, using the same five splits across all methods evaluated. We standardized the data using the mean and standard deviation calculated from the target class's training data. We report the average test performance across these five splits. During training, we used 5-fold cross-validation to select hyperparameters that achieved the highest evaluation score. 

A comprehensive set of evaluating metrics is reported over the test set to compare different OCC models. Accuracy (Accu) provides the ratio of correctly classified instances to the total number of instances, True Positive Rate (tpr) represents the proportion of positive instances correctly classified, while True Negative Rate (tnr) indicates the ratio of true negatives to the total number of negative samples. Precision (Pre) measures the proportion of instances classified as positive that are truly positive, and the F1-score (F1) is defined as the harmonic mean of precision and tpr. Additionally, Geometric Mean (GM) is employed to discern the best-performing parameters on the training set, calculated as the square root of the product of tpr and tnr. We selected hyperparameters based on the Geometric Mean, defined as: $ GM=\sqrt {tpr \times tnr},$ where $tpr$ is the true positive rate and $tnr$ is the true negative rate. The hyperparameter ranges were $C \in \{0.1, 0.2, 0.3, 0.4, 0.5\}$, $d \in \{1, 2, 3, 4, 5, 10, 5\}$, $\beta \in \{10^{-2}, 10^{-1}, 10^{0}, 10^{1}, 10^{2}\}$, and $\sigma \in \{10^{-1}, 10^{0}, 10^{1}, 10^{2}, 10^{3}\}$. For non-linear data descriptions, we applied the non-linear projection trick (NPT) \cite{kwak2013nonlinear}, with 
$\sigma$ determining the Gaussian kernel width. NPT is equivalent to the well-known kernel trick, but it allows the use of the linear variant of the method. For the gradient and Newton-based S-SVDD methods, we set a fixed learning rate (step size) of 0.1, with a total of 10 iterations.

\section{Data Analytics}
\label{sec:dataanalytics}
The exploratory data analysis focused on the following: identifying the most common keywords and hashtags found in the dataset, overall sentiment of the tweets as well as country-specific sentiment. The methods used were Word Cloud Display and Sentiment Analysis.  


Word cloud is a visualization method that displays how frequently certain keywords appear in a text by making the size of each word proportional to its frequency. As a first step,  we used it to analyze the most common keywords and hashtags in our dataset. 
The most popular hashtags in the harvested tweets are \texttt{`COVID-19', `lockdown', `vaccine', `CovidVaccine'} etc. 
The word `covid' is the most common, accounting for nearly~175,000 of all tweets. The next most common word is \texttt{`vaccin'} followed by \texttt{`lockdown', `stayhom'} and \texttt{`covidvaccin'}.
Next, we annotated the tweets based on their sentiments into two categories -- positive and negative. 
Words like \texttt{`live', `great', `good', `safe', `love'} were some of the words associated with positive tweets, 
while, words like \texttt{`difficult', `bad', `hard', `sick',} etc. are linked to the negative tweets category. 
In both word clouds words \texttt{`covid', `vaccine', `lockdown'} have the largest frequency of occurrence which makes the positive/negative tweet discussion challenging and inconclusive at the first level. 
 

%

\begin{figure*}[h]
	\begin{subfigure}{.5\textwidth}
		\centering
		\includegraphics[width=\linewidth]{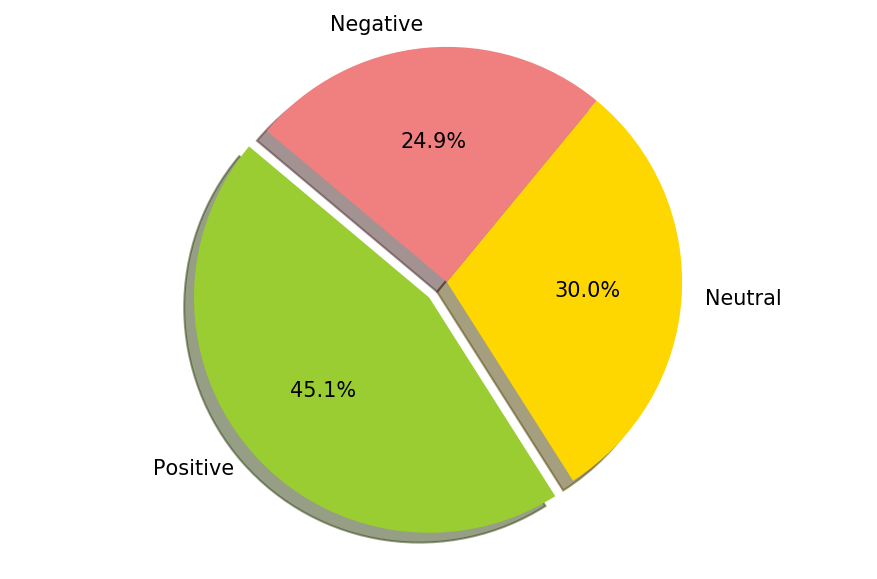}
		\caption{Overall sentiment analysis of tweets}
		\label{fig:overallsentiment}
	\end{subfigure}
	\begin{subfigure}{.5\textwidth}
		\centering
		\includegraphics[width=\linewidth]{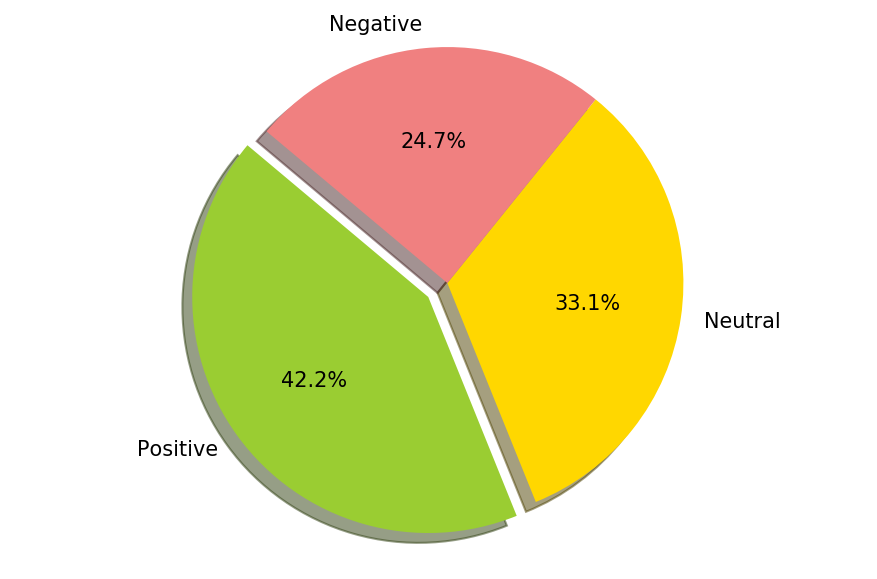}
		\caption{Sentiment analysis of vaccine-related tweets}
		\label{fig:sentimentvaccine}
	\end{subfigure}
	\caption{Sentiment analysis of tweets}
	\label{fig:sentimentanalysis}
\end{figure*}

The underlying $\mathbb{X}$ dataset was first subjected to sentiment analysis in order to better understand how people feel about the COVID-19 vaccine globally. We used the VADER sentiment analysis method to detect the polarity of the tweets. The polarity of a tweet is used to evaluate the emotions conveyed in the tweet. It can be either positive, negative, or neutral depending on the assigned score i.e., numerical values ranging from~-1 to~+1. Neutral sentiments are characterized by a polarity of exactly~0, and positive and negative emotions are indicated by a polarity greater than or less than~0, respectively.

The results of our analysis demonstrated that of all the COVID-19 tweets harvested,~45.1\% expressed positive overall sentiment,~30\% manifested a negative, and~24.9\% a neutral one. The distribution of the tweets' sentiment is given in Figure~\ref{fig:overallsentiment}. Our interpretation of these findings is that~45.1\% of all the tweets collected have shown a positive and confident attitude towards the virus. Subsequently, in order to classify the tweets into the vaccine `supporters' or `deniers' group, we filtered the data for just the COVID-19 vaccination-related tweets. The resulting datasets contained approximately~0.2 million tweets, out of which the positive sentiments were the highest~42.2\%, followed by neutral comprising~33.1\% of the tweets and finally, the negative sentiments accounted for~24.7\%, (see Figure~\ref{fig:sentimentvaccine}). The results showed that globally positive sentiments towards the vaccine are predominant and also in line with the positive attitude towards the virus. The same analogy was valid for 
neutral and negative sentiments towards the virus and vaccination

When looking at the country-specific tweets, we focused on the five countries with the most tweets, 
-- the United Kingdom, followed by the United States, India, Canada and South Africa. The United Kingdom, the United States, and India are some of the world's most affected countries, with a high number of COVID-19 cases. As a result, the high volume of COVID-19 related tweets from these countries is not surprising. A similar pattern was observed for vaccine-related tweets on a country level. 
However, the overall tweets' sentiments from these five countries were varied. For example, even though the United Kingdom is one of the most affected countries, the tweets' sentiments were more positive compared to India, the United States, and South Africa. In the United States, the positive and negative sentiments were almost equal, while the tweets' sentiments from India and South Africa leaned more towards the negative. 
Interestingly, for these countries, the analysis of the tweets' sentiments specific to vaccines leaned more towards the negative with the exception of South Africa.

Globally, the analysis showed an overwhelmingly positive tone towards the vaccine. However, contrasting results were obtained when the top five countries with the most tweets were looked at. Among the five countries, only tweets from South Africa exhibited a positive tone. These results were contrary to our expectations of having the most support in the most affected countries. 
\begin{table*}[!ht]
\caption{Performance metrics and standard deviations (S-) for positive tweets}
\label{table:positive-tweets}
\centering\resizebox{\textwidth}{!}{
\begin{tabular}{|l|c|c|c|c|c|c|c|c|c|c|c|c|}
\hline
\rowcolor[HTML]{D3D3D3} \textbf{Method} & \textbf{Accu} & \textbf{tpr} & \textbf{tnr} & \textbf{Pre} & \textbf{F1} & \textbf{GM} & \textbf{S-Accu} & \textbf{S-tpr} & \textbf{S-tnr} & \textbf{S-Pre} & \textbf{S-F1} & \textbf{S-GM} \\ \hline
\rowcolor{gray!10} \multicolumn{13}{|c|}{\textbf{LINEAR OCC}} \\
\hline
S-SVDD$\psi1$-min & 0.77 & 0.83 & 0.71 & 0.75 & 0.79 & 0.76 & 0.09 & 0.08 & 0.15 & 0.11 & 0.07 & 0.09 \\ \hline
S-SVDD$\psi2$-min & 0.78 & 0.82 & 0.73 & 0.76 & 0.79 & 0.77 & 0.08 & 0.07 & 0.13 & 0.10 & 0.07 & 0.08 \\ \hline
S-SVDD$\psi3$-min & 0.75 & 0.83 & 0.67 & 0.73 & 0.77 & 0.74 & 0.11 & 0.11 & 0.19 & 0.13 & 0.10 & 0.12 \\ \hline
S-SVDD$\psi4$-min & 0.76 & 0.81 & 0.72 & 0.75 & 0.78 & 0.76 & 0.09 & 0.06 & 0.14 & 0.11 & 0.08 & 0.09 \\ \hline
S-SVDD$\psi1$-max & 0.77 & 0.84 & 0.69 & 0.74 & 0.78 & 0.76 & 0.09 & 0.06 & 0.14 & 0.11 & 0.08 & 0.10 \\ \hline
S-SVDD$\psi2$-max & 0.78 & 0.82 & 0.73 & 0.76 & 0.79 & 0.77 & 0.08 & 0.07 & 0.13 & 0.10 & 0.07 & 0.08 \\ \hline
S-SVDD$\psi3$-max & 0.77 & 0.84 & 0.69 & 0.74 & 0.78 & 0.76 & 0.09 & 0.08 & 0.14 & 0.11 & 0.08 & 0.10 \\ \hline
S-SVDD$\psi4$-max & 0.77 & 0.83 & 0.71 & 0.75 & 0.79 & 0.77 & 0.09 & 0.06 & 0.13 & 0.10 & 0.08 & 0.10 \\ \hline
NS-SVDD$\psi1$-min & 0.78 & 0.82 & 0.73 & 0.76 & 0.79 & 0.77 & 0.08 & 0.07 & 0.13 & 0.10 & 0.07 & 0.08 \\ \hline
NS-SVDD$\psi2$-min & 0.80 & 0.77 & 0.83 & 0.83 & 0.79 & 0.79 & 0.07 & 0.16 & 0.10 & 0.07 & 0.09 & 0.08 \\ \hline
NS-SVDD$\psi3$-min & 0.77 & 0.81 & 0.73 & 0.76 & 0.78 & 0.77 & 0.09 & 0.08 & 0.13 & 0.10 & 0.08 & 0.09 \\ \hline
NS-SVDD$\psi4$-min & 0.76 & 0.81 & 0.71 & 0.75 & 0.77 & 0.75 & 0.12 & 0.07 & 0.19 & 0.13 & 0.09 & 0.13 \\ \hline
NS-SVDD$\psi1$-max & 0.78 & 0.82 & 0.73 & 0.76 & 0.79 & 0.77 & 0.08 & 0.07 & 0.13 & 0.10 & 0.07 & 0.08 \\ \hline
NS-SVDD$\psi2$-max & 0.81 & 0.82 & 0.81 & 0.81 & 0.81 & 0.81 & 0.04 & 0.08 & 0.08 & 0.06 & 0.05 & 0.05 \\ \hline
NS-SVDD$\psi3$-max & 0.78 & 0.82 & 0.73 & 0.76 & 0.79 & 0.77 & 0.08 & 0.07 & 0.13 & 0.10 & 0.07 & 0.08 \\ \hline
NS-SVDD$\psi4$-max & 0.74 & 0.83 & 0.64 & 0.72 & 0.76 & 0.72 & 0.11 & 0.06 & 0.23 & 0.11 & 0.07 & 0.14 \\ \hline
OCSVM & 0.46 & 0.63 & 0.29 & 0.47 & 0.54 & 0.41 & 0.09 & 0.04 & 0.16 & 0.07 & 0.05 & 0.11 \\ \hline
SVDD & 0.77 & 0.81 & 0.73 & 0.76 & 0.78 & 0.77 & 0.10 & 0.09 & 0.13 & 0.11 & 0.09 & 0.10 \\ \hline
ESVDD & 0.73 & 0.69 & 0.77 & 0.76 & 0.72 & 0.73 & 0.07 & 0.08 & 0.12 & 0.09 & 0.06 & 0.07 \\ \hline
\rowcolor{gray!10} \multicolumn{13}{|c|}{\textbf{Non-LINEAR OCC}} \\
\hline

S-SVDD$\psi1$-min & 0.78 & 0.78 & 0.78 & 0.78 & 0.78 & 0.78 & 0.05 & 0.08 & 0.08 & 0.06 & 0.05 & 0.05 \\ \hline
S-SVDD$\psi2$-min & 0.78 & 0.83 & 0.73 & 0.76 & 0.79 & 0.78 & 0.09 & 0.10 & 0.13 & 0.10 & 0.09 & 0.10 \\ \hline
S-SVDD$\psi3$-min & 0.79 & 0.79 & 0.78 & 0.79 & 0.79 & 0.78 & 0.03 & 0.10 & 0.07 & 0.05 & 0.04 & 0.03 \\ \hline
S-SVDD$\psi4$-min & 0.78 & 0.78 & 0.79 & 0.79 & 0.78 & 0.78 & 0.04 & 0.11 & 0.07 & 0.05 & 0.05 & 0.04 \\ \hline
S-SVDD$\psi1$-max & 0.79 & 0.77 & 0.81 & 0.81 & 0.77 & 0.78 & 0.08 & 0.20 & 0.11 & 0.07 & 0.11 & 0.09 \\ \hline
S-SVDD$\psi2$-max & 0.77 & 0.73 & 0.80 & 0.79 & 0.75 & 0.76 & 0.09 & 0.16 & 0.11 & 0.09 & 0.11 & 0.10 \\ \hline
S-SVDD$\psi3$-max & 0.75 & 0.69 & 0.81 & 0.81 & 0.72 & 0.73 & 0.07 & 0.23 & 0.13 & 0.08 & 0.14 & 0.10 \\ \hline
S-SVDD$\psi4$-max & 0.78 & 0.76 & 0.81 & 0.81 & 0.77 & 0.77 & 0.07 & 0.19 & 0.11 & 0.06 & 0.11 & 0.09 \\ \hline
NS-SVDD$\psi1$-min & 0.77 & 0.81 & 0.73 & 0.76 & 0.78 & 0.77 & 0.07 & 0.05 & 0.12 & 0.09 & 0.06 & 0.08 \\ \hline
NS-SVDD$\psi2$-min & 0.78 & 0.85 & 0.71 & 0.76 & 0.80 & 0.77 & 0.06 & 0.09 & 0.17 & 0.10 & 0.05 & 0.07 \\ \hline
NS-SVDD$\psi3$-min & 0.72 & 0.79 & 0.65 & 0.74 & 0.75 & 0.66 & 0.13 & 0.11 & 0.35 & 0.14 & 0.06 & 0.27 \\ \hline
NS-SVDD$\psi4$-min & 0.79 & 0.83 & 0.76 & 0.78 & 0.80 & 0.79 & 0.07 & 0.08 & 0.08 & 0.07 & 0.07 & 0.07 \\ \hline
NS-SVDD$\psi1$-max & 0.77 & 0.73 & 0.81 & 0.80 & 0.75 & 0.76 & 0.09 & 0.22 & 0.08 & 0.06 & 0.15 & 0.11 \\ \hline
NS-SVDD$\psi2$-max & 0.79 & 0.79 & 0.79 & 0.80 & 0.79 & 0.79 & 0.05 & 0.05 & 0.08 & 0.06 & 0.04 & 0.04 \\ \hline
NS-SVDD$\psi3$-max & 0.76 & 0.69 & 0.82 & 0.80 & 0.73 & 0.74 & 0.07 & 0.19 & 0.09 & 0.06 & 0.12 & 0.09 \\ \hline
NS-SVDD$\psi4$-max & 0.77 & 0.72 & 0.82 & 0.81 & 0.74 & 0.76 & 0.08 & 0.21 & 0.09 & 0.06 & 0.14 & 0.11 \\ \hline
OCSVM & 0.53 & 0.43 & 0.63 & 0.60 & 0.45 & 0.43 & 0.16 & 0.24 & 0.40 & 0.22 & 0.20 & 0.20 \\ \hline
SVDD & 0.78 & 0.81 & 0.74 & 0.76 & 0.79 & 0.77 & 0.06 & 0.07 & 0.11 & 0.08 & 0.05 & 0.06 \\ \hline
ESVDD & 0.74 & 0.69 & 0.79 & 0.77 & 0.72 & 0.73 & 0.06 & 0.11 & 0.10 & 0.07 & 0.08 & 0.07 \\
\hline
\end{tabular}
}
\end{table*}

\textbf{OCC-Based Results and Discussion} From the collected tweets, we selected~200 tweets, with~100 tweets manually labeled as positive and~100 as negative by human experts. In~\autoref{table:positive-tweets}, we report the average performance measures of various OCC methods on five data splits of the positive tweets. The classifiers are grouped into two categories: linear OCC and non-linear OCC. In ~\autoref{table:positive-tweets}, we also report the STD of evaluating metrics for the linear and non-linear methods over the five data splits of the dataset. Similarly, in~\autoref{table:negative-tweets}, we present the average performance and standard deviation of various OCC methods for the five data splits, but applied to the negative tweets.

Considering the GM values, NS-SVDD$\psi2$-max demonstrates superior performance compared to the other linear OCC. This trend is observed for both positive~\autoref{table:positive-tweets} and negative~\autoref{table:negative-tweets} tweets. For positive tweets, NS-SVDD$\psi2$-max outperforms all other methods by achieving a GM value of~0.81. This is likely due to its ability to focus on maximizing the variance captured in the regularization term $\psi$, which may better represent the characteristics of the positive class. For the negative tweets, NS-SVDD$\psi1$-min gives the best performance among the OCC methods. Its superior performance can be attributed to its specific handling of the regularization term $\psi$. By minimizing the variance within the project feature space, it achieves a more precise description of the negative class, leading to better classification outcomes. Among the non-linear OCC classifiers, for the positive tweets, both NS-SVDD$\psi4$-min and NS-SVDD$\psi2$-max achieve the highest GM value of~0.79. While, for the non-negative tweets, S-SVDD$\psi1$-min achieves the best GM value of 0.79.

\begin{table*}[!ht]
\centering
\caption{Performance metrics and standard deviations (S-) for negative tweets}
\label{table:negative-tweets}
\resizebox{\textwidth}{!}{
\begin{tabular}{|l|c|c|c|c|c|c|c|c|c|c|c|c|c|}
\hline
\rowcolor{gray!20} \textbf{Method} & \textbf{Accu} & \textbf{tpr} & \textbf{tnr} & \textbf{Pre} & \textbf{F1} & \textbf{GM} & \textbf{S-Accu} & \textbf{S-tpr} & \textbf{S-tnr} & \textbf{S-Pre} & \textbf{S-F1} & \textbf{S-GM} \\
\hline
\rowcolor{gray!10} \multicolumn{13}{|c|}{\textbf{LINEAR OCC}} \\
\hline
S-SVDD$\psi1$-min & 0.72 & 0.81 & 0.63 & 0.69 & 0.74 & 0.71 & 0.08 & 0.10 & 0.13 & 0.09 & 0.08 & 0.09 \\ \hline
S-SVDD$\psi2$-min & 0.73 & 0.83 & 0.63 & 0.69 & 0.75 & 0.72 & 0.08 & 0.09 & 0.09 & 0.07 & 0.07 & 0.08 \\ \hline
S-SVDD$\psi3$-min & 0.70 & 0.77 & 0.63 & 0.68 & 0.72 & 0.69 & 0.08 & 0.11 & 0.13 & 0.08 & 0.07 & 0.08 \\ \hline
S-SVDD$\psi4$-min & 0.72 & 0.76 & 0.67 & 0.70 & 0.73 & 0.71 & 0.07 & 0.13 & 0.09 & 0.07 & 0.08 & 0.07 \\ \hline
S-SVDD$\psi1$-max & 0.73 & 0.84 & 0.63 & 0.69 & 0.76 & 0.72 & 0.06 & 0.10 & 0.10 & 0.06 & 0.06 & 0.07 \\ \hline
S-SVDD$\psi2$-max & 0.72 & 0.85 & 0.59 & 0.68 & 0.75 & 0.70 & 0.11 & 0.07 & 0.17 & 0.09 & 0.08 & 0.13 \\ \hline
S-SVDD$\psi3$-max & 0.73 & 0.83 & 0.63 & 0.69 & 0.75 & 0.72 & 0.08 & 0.10 & 0.07 & 0.07 & 0.08 & 0.08 \\ \hline
S-SVDD$\psi4$-max & 0.74 & 0.85 & 0.63 & 0.70 & 0.76 & 0.72 & 0.06 & 0.09 & 0.10 & 0.05 & 0.06 & 0.07 \\ \hline
NS-SVDD$\psi1$-min & 0.80 & 0.82 & 0.77 & 0.79 & 0.80 & 0.79 & 0.04 & 0.08 & 0.08 & 0.05 & 0.05 & 0.04 \\ \hline
NS-SVDD$\psi2$-min & 0.74 & 0.78 & 0.69 & 0.74 & 0.75 & 0.72 & 0.07 & 0.15 & 0.20 & 0.11 & 0.07 & 0.08 \\ \hline
NS-SVDD$\psi3$-min & 0.74 & 0.81 & 0.67 & 0.72 & 0.76 & 0.73 & 0.09 & 0.06 & 0.15 & 0.10 & 0.08 & 0.10 \\ \hline
NS-SVDD$\psi4$-min & 0.75 & 0.87 & 0.62 & 0.70 & 0.78 & 0.73 & 0.10 & 0.10 & 0.15 & 0.09 & 0.08 & 0.11 \\ \hline
NS-SVDD$\psi1$-max & 0.72 & 0.85 & 0.59 & 0.68 & 0.76 & 0.70 & 0.10 & 0.06 & 0.17 & 0.09 & 0.07 & 0.12 \\ \hline
NS-SVDD$\psi2$-max & 0.73 & 0.85 & 0.61 & 0.69 & 0.76 & 0.71 & 0.09 & 0.07 & 0.13 & 0.08 & 0.07 & 0.10 \\ \hline
NS-SVDD$\psi3$-max & 0.73 & 0.87 & 0.60 & 0.69 & 0.77 & 0.71 & 0.11 & 0.05 & 0.18 & 0.10 & 0.08 & 0.13 \\ \hline
NS-SVDD$\psi4$-max & 0.72 & 0.86 & 0.59 & 0.69 & 0.76 & 0.70 & 0.11 & 0.06 & 0.19 & 0.11 & 0.08 & 0.13 \\ \hline
OCSVM & 0.48 & 0.55 & 0.41 & 0.53 & 0.52 & 0.42 & 0.15 & 0.13 & 0.34 & 0.16 & 0.08 & 0.19 \\ \hline
SVDD & 0.75 & 0.81 & 0.68 & 0.72 & 0.76 & 0.74 & 0.07 & 0.09 & 0.08 & 0.07 & 0.07 & 0.07 \\ \hline
ESVDD & 0.75 & 0.73 & 0.78 & 0.78 & 0.75 & 0.75 & 0.05 & 0.10 & 0.11 & 0.10 & 0.06 & 0.05 \\ \hline
\rowcolor{gray!10} \multicolumn{13}{|c|}{\textbf{Non-LINEAR OCC}} \\
\hline
S-SVDD$\psi1$-min & 0.80 & 0.78 & 0.81 & 0.81 & 0.79 & 0.79 & 0.04 & 0.10 & 0.08 & 0.06 & 0.05 & 0.04 \\ \hline
S-SVDD$\psi2$-min & 0.75 & 0.77 & 0.73 & 0.74 & 0.75 & 0.75 & 0.08 & 0.13 & 0.05 & 0.06 & 0.09 & 0.08 \\ \hline
S-SVDD$\psi3$-min & 0.77 & 0.76 & 0.77 & 0.79 & 0.76 & 0.76 & 0.08 & 0.11 & 0.16 & 0.12 & 0.07 & 0.08 \\ \hline
S-SVDD$\psi4$-min & 0.77 & 0.77 & 0.77 & 0.79 & 0.77 & 0.77 & 0.09 & 0.11 & 0.15 & 0.11 & 0.07 & 0.09 \\ \hline
S-SVDD$\psi1$-max & 0.73 & 0.82 & 0.65 & 0.70 & 0.75 & 0.73 & 0.06 & 0.10 & 0.06 & 0.05 & 0.06 & 0.06 \\ \hline
S-SVDD$\psi2$-max & 0.73 & 0.81 & 0.65 & 0.70 & 0.75 & 0.73 & 0.05 & 0.10 & 0.05 & 0.04 & 0.06 & 0.05 \\ \hline
S-SVDD$\psi3$-max & 0.75 & 0.81 & 0.68 & 0.72 & 0.76 & 0.74 & 0.07 & 0.13 & 0.06 & 0.05 & 0.09 & 0.07 \\ \hline
S-SVDD$\psi4$-max & 0.74 & 0.85 & 0.64 & 0.70 & 0.77 & 0.73 & 0.08 & 0.07 & 0.11 & 0.07 & 0.07 & 0.09 \\ \hline
NS-SVDD$\psi1$-min & 0.71 & 0.81 & 0.61 & 0.68 & 0.73 & 0.70 & 0.07 & 0.13 & 0.10 & 0.06 & 0.08 & 0.07 \\ \hline
NS-SVDD$\psi2$-min & 0.75 & 0.75 & 0.75 & 0.76 & 0.75 & 0.74 & 0.07 & 0.11 & 0.15 & 0.12 & 0.07 & 0.07 \\ \hline
NS-SVDD$\psi3$-min & 0.77 & 0.83 & 0.71 & 0.75 & 0.78 & 0.76 & 0.06 & 0.10 & 0.12 & 0.09 & 0.06 & 0.06 \\ \hline
NS-SVDD$\psi4$-min & 0.77 & 0.83 & 0.71 & 0.74 & 0.78 & 0.77 & 0.06 & 0.07 & 0.04 & 0.04 & 0.06 & 0.06 \\ \hline
NS-SVDD$\psi1$-max & 0.74 & 0.83 & 0.65 & 0.70 & 0.76 & 0.73 & 0.05 & 0.08 & 0.06 & 0.04 & 0.05 & 0.05 \\ \hline
NS-SVDD$\psi2$-max & 0.77 & 0.77 & 0.78 & 0.80 & 0.77 & 0.77 & 0.08 & 0.10 & 0.16 & 0.14 & 0.07 & 0.08 \\ \hline
NS-SVDD$\psi3$-max & 0.74 & 0.81 & 0.67 & 0.71 & 0.75 & 0.74 & 0.08 & 0.13 & 0.04 & 0.05 & 0.09 & 0.07 \\ \hline
NS-SVDD$\psi4$-max & 0.74 & 0.80 & 0.69 & 0.72 & 0.76 & 0.74 & 0.08 & 0.10 & 0.08 & 0.07 & 0.08 & 0.08 \\ \hline
OCSVM & 0.45 & 0.58 & 0.33 & 0.48 & 0.52 & 0.40 & 0.17 & 0.16 & 0.28 & 0.15 & 0.13 & 0.19 \\ \hline
SVDD & 0.75 & 0.81 & 0.69 & 0.73 & 0.77 & 0.75 & 0.07 & 0.09 & 0.09 & 0.06 & 0.07 & 0.07 \\ \hline
ESVDD & 0.73 & 0.63 & 0.83 & 0.81 & 0.70 & 0.72 & 0.06 & 0.10 & 0.12 & 0.12 & 0.06 & 0.06 \\ \hline

\end{tabular}}
\end{table*}

These results indicate that choice of regularization term usage $\psi$ and optimization approach (min vs max) significantly influence the performance of the S-SVDD variants. By refining the projection matrix $\mathbf{Q}$ through newton-based optimization, NS-SVDD$\psi$-min and NS-SVDD$\psi$-max effectively adapt to the specific characteristics of the positive and negative tweets dataset, respectively. Other OCCs, such as SVDD and ESVDD, demonstrate average performance. While their performance is better than those of OCSVM, they fall short when compared to the performance of S-SVDD variants. This difference can be attributed to the fact that S-SVDD incorporates a regularization term ($\psi$) that effectively captures class variance and adapts to the data distribution, providing a more refined boundary. In contrast, SVDD and ESVDD lack this enhanced regularization mechanism, which limits their ability to capture the underlying data structure as effectively as S-SVDD. However, their simpler formulations and reliance on a standard optimization still enable them to outperform OCSVM.

\section{Conclusion}
\label{sec:conclusion}

Vaccines provide the best chance of success in the fight against COVID-19. The unprecedented success and speed in making them available, however, brought about a major challenge in the form of vaccine hesitancy. Both sides of the spectrum -- vaccine supporters and vaccine deniers -- are present and supported in social media. Assessing their opinions and tweaking the approaches used globally and locally can make a significant difference to the final outcome of vaccine campaigns.  
In an effort to offer a deeper understanding of this problem, we examined the public sentiment towards the virus and the vaccine. We created a dataset consisting of~0.4 million tweets related to COVID-19 and COVID-19 vaccine and performed textual and sentiment analysis in order to gauge public perception towards both. The results showed that the overall global sentiment was mainly positive. The analysis at the country level, however, yielded opposite results. The countries most severely hit by the pandemic still have a negative public perception towards vaccines. Furthermore, we collected a set of~200 tweets, with~100 manually labeled as positive and~100 as negative by human expert. These labeled samples were used to train different OCCs, and results demonstrated that S-SVDD achieved superior performance compared to other OCCs.

We believe our work can be used as a tool on both strategic and operational levels when it comes to dealing with the problem of vaccine hesitancy. On a strategic level, the form of providing insights for policymakers in order to devise more effective vaccine delivery strategies on a both global and local levels, and on the operational, to ensure timely management and smooth roll-out of specific vaccine campaigns.

\section*{Acknowledgment}
This work was funded by the SWARMCHESTRATE EU research project (No. 101135012).


\bibliographystyle{splncs04}
\bibliography{mybibliography}

\appendix

\end{document}